\documentclass{article}

\PassOptionsToPackage{numbers, compress}{natbib}
\usepackage[preprint]{neurips_2025}

\usepackage[utf8]{inputenc} 
\usepackage[T1]{fontenc}    
\usepackage{hyperref}       
\usepackage{url}            
\usepackage{booktabs}       
\usepackage{amsfonts}       
\usepackage{nicefrac}       
\usepackage{microtype}      
\usepackage{xcolor}         

\usepackage{algorithmicx}
\usepackage{algpseudocode}
\usepackage{subcaption}
\usepackage{amsmath,xcolor}
\usepackage{graphicx}
\usepackage{enumitem}
\algrenewcommand\algorithmicindent{1.2em}
\algrenewcommand\alglinenumber[1]{\footnotesize #1}
\usepackage{xspace}
\newcommand{\name}{MambaLite-Micro\xspace}
\title{MambaLite-Micro: Memory-Optimized Mamba Inference on MCUs}
\workshoptitle{ML For Systems}

\author{%
  Hongjun Xu,\quad  Junxi Xia,\quad  Weisi Yang,\quad  Yueyuan Sui,\quad  Stephen Xia\\
  Northwestern University, Evanston, IL\\
  \texttt{\{r2f9a9, JunxiXia2024, weisiyang2029, yueyuansui2024\}@u.northwestern.edu, }\\
  \texttt{stephen.xia@northwestern.edu} \\
}

\begin{document}

\maketitle

\begin{abstract}
Deploying Mamba models on microcontrollers (MCUs) remains challenging due to limited memory, the lack of native operator support, and the absence of embedded-friendly toolchains. \textit{We present, to our knowledge, the first deployment of a Mamba-based neural architecture on a resource-constrained MCU}, a fully C-based runtime-free inference engine: \name. Our pipeline maps a trained PyTorch Mamba model to on-device execution by (1) exporting model weights into a lightweight format, and (2) implementing a handcrafted Mamba layer and supporting operators in C with operator fusion and memory layout optimization. \name  eliminates large intermediate tensors, reducing \textbf{83.0\%} peak memory, while maintaining an average numerical error of only \(1.7\times 10^{-5}\) relative to the PyTorch Mamba implementation. When evaluated on keyword spotting (KWS) and human activity recognition (HAR) tasks, \name achieved \textbf{100\% consistency} with the PyTorch baselines, fully preserving classification accuracy. We further validated portability by deploying on both ESP32S3 and STM32H7 microcontrollers, demonstrating consistent operation across heterogeneous embedded platforms and paving the way for bringing advanced sequence models like Mamba to real-world resource-constrained applications.

\end{abstract}

\section{Introduction}
The rapid development of deep learning has brought increasingly complex models to a wide range of applications, including those on resource-constrained embedded devices~\cite{sandler2018mobilenetv2,han2015deep,banbury2020benchmarking}. However, deploying modern sequence models on MCUs remains challenging due to severe limitations in memory capacity, compute throughput, and power budget~\cite{lin2020mcunet,david2021tensorflow}. While architectures such as Transformers and LSTMs have achieved remarkable success in sequence modeling, their memory footprint and computational complexity often limit their MCU deployments~\cite{9556405,9912325}.

The recently proposed Mamba architecture offers an attractive alternative for embedded scenarios~\cite{gu2023mamba}. By using selective state-space modeling, Mamba carefully compresses the latent state space by selectively retaining and forgetting parts of the history to achieve more powerful representations than LSTMs, while being more efficient than transformer architectures~\cite{gu2023mamba,dao2024transformers}. However, its reference implementation relies on custom Triton GPU kernels and lacks an ONNX export path, making it incompatible with standard embedded AI toolchains and hindering direct deployment to MCUs. Moreover, previous “embedded” Mamba experiments have typically stopped short of actual MCU execution, instead relying on desktop inference or simulation, leaving true on-device feasibility unproven~\cite{suwannaphong2025optimising}.

In this work, we address these challenges by presenting, to our knowledge, the first complete workflow for deploying a PyTorch Mamba model directly onto embedded hardware. Our approach is built around a fully C-based runtime-free inference engine \name, which enables portability across heterogeneous MCU and embedded platforms without reliance on vendor-specific runtimes. Specifically, we (1) export the trained PyTorch model into a lightweight weight format, (2) implement a handcrafted Mamba layer and supporting operators in C, and (3) apply operator fusion and memory layout optimization to reduce runtime overhead while preserving precision. \name reproduces the PyTorch implementation with an average error of only \(1.7\times 10^{-5}\), while downstream classification tasks achieve \textbf{100\% consistency} with PyTorch baselines. Validated on ESP32S3 and STM32H7 MCUs, our results demonstrate that advanced sequence modeling is feasible on highly resource-constrained devices, paving the way for real-world embedded applications of Mamba. The code will be released at \href{https://github.com/Whiten-Rock/MambaLite-Micro}{github.com/Whiten-Rock/MambaLite-Micro}.

\begin{figure}[ht]
    \centering
    \includegraphics[width=\linewidth]{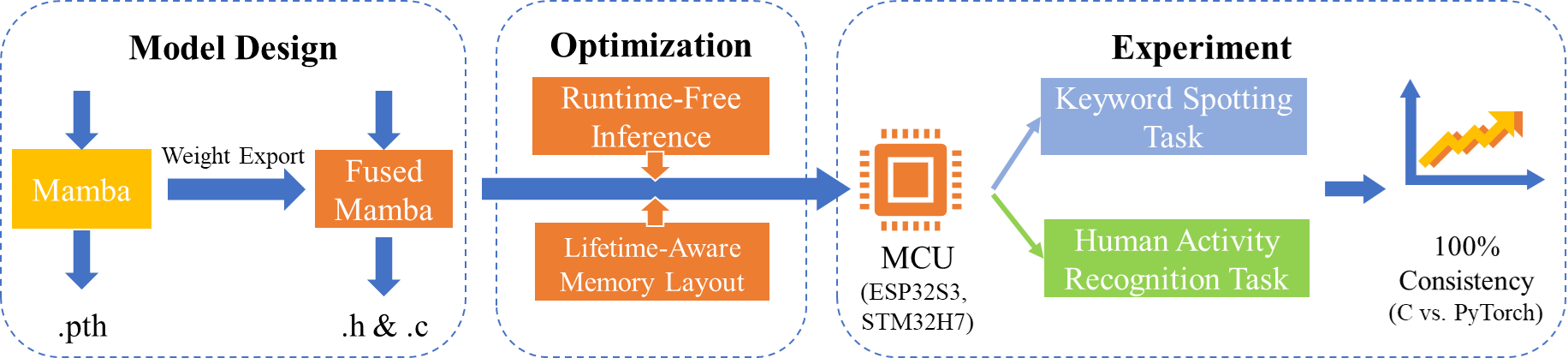}
    \caption{End-to-end deployment pipeline of \name. A trained PyTorch Mamba model is exported and mapped to a fully C-based inference engine through operator fusion and lifetime-aware memory layout optimization. This runtime-free design eliminates large intermediate tensors and significantly reduces memory footprint. We demonstrate our baseline can maintain \textbf{100\% consistency} with PyTorch baselines on diverse application domains, including keyword spotting and human activity recognition.}
    \label{fig:method}
\end{figure}

\section{Methodology}
We present the overall deployment pipeline of \name (Fig.~\ref{fig:method}), which consists of runtime-free inference, operator fusion, and lifetime-aware memory layout management.
\subsection{Runtime-free Embedded Mamba Inference}
To maximize portability on microcontrollers, we reimplement the PyTorch Mamba inference logic entirely in~C, faithfully reproducing the computational flow of the reference model. Unlike the original implementation, which relies on custom Triton GPU kernels and lacks ONNX compatibility, our design, \name , requires no extra runtime. Instead, the trained PyTorch weights are converted into plain~C arrays and compiled directly into the MCU firmware, yielding a self-contained binary executable, as illustrated in the runtime-free inference stage of Fig.~\ref{fig:method}. 

\subsection{Operator Fusion in the Mamba Computation}
As highlighted in Fig.~\ref{fig:method}, operator fusion is a key component in \name to eliminate large intermediate tensors and reduce memory.

\begin{figure*}[ht]
\centering
\begin{minipage}[t]{0.48\linewidth}
\footnotesize
\textbf{Algorithm 1\quad Original Mamba (Alg.~2, Steps 1--6)}\\[-2pt]
\rule{\linewidth}{0.4pt}
\begin{algorithmic}[1]
\Statex \textbf{Input:} $x :(B, L, D)$
\Statex \textbf{Output:} $y :(B, L, D)$
\State $A :(D,N) \gets \text{Parameter}$
\State $B :(B, L, N) \gets s_B(x)$
\State $C :(B, L, N) \gets s_C(x)$
\State $\Delta :(B, L, D) \gets \tau_{\Delta}(\text{Param}+s_{\Delta}(x))$
\State \textcolor{red}{$\bar A,\ \bar B :(B,L, D, N)$} $\gets \text{discretize}(\Delta, A, B)$
\State $y \gets \text{SSM}(\bar A, \bar B, C)(x)$
\State \Return $y$
\end{algorithmic}
\rule{\linewidth}{0.4pt}
\end{minipage}\hfill
\begin{minipage}[t]{0.48\linewidth}
\footnotesize
\textbf{Algorithm 2\quad Proposed Fused Implementation}\\[-2pt]
\rule{\linewidth}{0.4pt}
\begin{algorithmic}[1]
\Statex \textbf{Input:} $x :(B, L, D)$
\Statex \textbf{Output:} $y :(B, L, D)$
\State $A :(D,N) \gets \text{Parameter}$
\State $B :(B, L, N) \gets s_B(x)$
\State $C :(B, L, N) \gets s_C(x)$
\State $\Delta :(B, L, D) \gets \tau_{\Delta}(\text{Param}+s_{\Delta}(x))$
\State $y \gets $\textcolor{red}{$\text{SSM}(A, B, C,\Delta)(x)$}
\State \Return $y$
\Statex
\end{algorithmic}
\rule{\linewidth}{0.4pt}
\end{minipage}

\vspace{2pt}
\label{fig:alg-compare}
\end{figure*}
In the original Mamba implementation, Algorithm~2 (``SSM + Selection'') Step~5 explicitly constructs the 4D tensors
\begin{equation}
    \bar A = \exp\!\big(\mathrm{einsum}(\texttt{'bdl,dn->bdln'}, \delta, A)\big), 
    \qquad
    \bar B_{u} = \mathrm{einsum}(\texttt{'bdl,dn,bdl->bdln'}, \delta, B, u),
\end{equation}
which requires $(B,D,L,N)$ storage~\cite{gu2023mamba}. At time step $i$, the recurrence update is
\begin{equation}
    x \;\leftarrow\; \bar A[:,:,i,:] \odot x \;+\; \bar B_{u}[:,:,i,:].
\end{equation}
This large intermediate footprint is prohibitive for MCU deployment.

\name fuses Step~5 into the recurrence of Step~6.
Instead of materializing $\bar A$ and $\bar B_{u}$, the recurrence can be computed on-the-fly:
\begin{equation}
    \bar A[:,:,i,:] = \exp\!\big(\delta[:,:,i] \odot A\big), 
    \qquad 
    \bar B_{u}[:,:,i,:] = \delta[:,:,i] \odot (B \odot u),
\end{equation}
so that each update reduces to
\begin{equation}
    x \;\leftarrow\; \exp(\delta[:,:,i]\odot A)\, x \;+\; \delta[:,:,i]\odot (B \odot u),
    \qquad
    y \;\leftarrow\; \mathrm{einsum}(\texttt{'bdn,dn->bd'}, x, C).
\end{equation}

This fusion eliminates the need for $(B,D,L,N)$ intermediates, reducing the memory requirement from 
$\mathcal{O}(BDLN)$ to $\mathcal{O}(BDN)$. Moreover, it enables streaming execution along the sequence length $L$, and thus significantly reducing peak RAM usage while preserving numerical accuracy.

\subsection{Lifetime-Aware Memory Layout Management}
We further reduce memory usage through lifetime-aware buffer allocation, as shown in Fig.~\ref{fig:method}, where intermediate buffers are created only for the duration of their use and memory is reused across non-overlapping lifetimes. Coupled with fused computation, this strategy minimizes both peak RAM requirements and runtime overhead. Overall, \name's runtime-free design—combining lifetime-aware allocation, operator fusion, and memory layout management—eliminates large intermediate tensors, reduces peak memory usage, and enables consistent execution across diverse embedded hardware.

\section{Experiments}
As shown in the Fig.~\ref{fig:method}, we evaluate \name using two representative downstream tasks: 
\textbf{keyword spotting (KWS)} and \textbf{human activity recognition (HAR)}. For KWS, we adopt the Speech Commands v2 dataset~\cite{speechcommandsv2}, 
while HAR is based on the UCI-HAR dataset~\cite{anguita2013public}. We further deploy on two representative MCU platforms, ESP32S3 and STM32H7, to validate its portability. The detailed preprocessing and experimental setup are provided in the Appendix \ref{experiment}.

\subsection{Evaluation Metrics}
We evaluate \name on numerical fidelity, classification consistency and system-level performance. 
At the layer level, numerical agreement is assessed using the average $L_\infty$ error (``max error''), the average mean error, and the worst-case $L_\infty$ error across samples over the Mamba layer outputs. At the end-to-end level, we evaluate classification consistency with respect to the PyTorch reference model, as well as inference latency and peak memory footprint (RAM) measured during runtime execution.

\section{Results \& Discussion}

\paragraph{Numerical fidelity}
Both implementations operate in fp32 precision. Across multiple tasks 
(3-class and 10-class KWS, and 6-class HAR), the MCU outputs closely 
match the PyTorch reference: average sample-level errors remain in the 
range of $10^{-5}{-}10^{-4}$, with worst-case deviations below 
$1.5{\times}10^{-3}$ (Table~\ref{tab:num}). Across all three 
tasks, the overall average error is only $1.7\times 10^{-5}$. Despite 
these small discrepancies, downstream classification accuracy is fully 
preserved.


\paragraph{Accuracy vs agreement.}
MCU predictions are \emph{identical} to the PyTorch baseline across both KWS and HAR tasks, 
demonstrating faithful porting with \textbf{100\% consistency}. 
The baseline accuracy and confusion matrix, which show 
the expected classification performance on the test set, 
are provided in the Appendix \ref{confusion matrix}.

\paragraph{Memory \& Latency}
Table~\ref{tab:perf} presents the ablation results on resource usage and latency. 
On the ESP32S3, the naive unfused baseline without lifetime management requires 
$1{,}384{,}472$\,B ($1352$\,KB) of peak RAM, 
which exceeds the available capacity of this device. 
With lifetime-aware buffering and operator fusion, the peak memory footprint 
is reduced to $235{,}620$\,B ($230$\,KB), achieving a \textbf{83.0\% reduction}.

For the KWS task, each input is a $4{,}000$-point mel-spectrogram preprocessed from an audio segment sampled at 16 kHz. Within the model, these features are linearly projected into a $64 \times 100$ representation that forms the input sequence to the Mamba layer. The inference of \name takes $1133.2$\,ms in fp32 on the ESP32S3.
On the STM32H7, the same workload completes in $934.9$\,ms. Notably, increasing the number of output classes from 3 to 10 in KWS has negligible impact on peak memory footprint, inference latency, or flash usage, since the computational overhead of the final classification layer is minimal compared to the sequence processing cost. These results demonstrate the portability and numerical fidelity of the \name across heterogeneous MCU platforms. 

Despite operating in full fp32 precision, the observed KWS 
throughput is already comparable to MCU deployments of attention-based models 
that have been quantized to int8. 
We benchmarked a TFLM-based attention model with int8 quantization as an reference for the same task. The attention layer is configured with 4 heads of dimension 16, giving the same 64 hidden dimension over a sequence length of 100 as \name, and which requires $3{,}991$ ms for inference. In contrast, the same attention-based model in fp32 could not be reliably 
benchmarked on the MCU, as the runtime was prohibitively long. 
This highlights that \name already achieves a substantial reduction in inference time compared to quantized attention baselines, while still leaving room for further optimizations such as post-training quantization, fixed-point arithmetic, or SIMD acceleration.

For the HAR task, on the ESP32S3, execution requires only 44{,}244\,B (43.2\,KB) of peak RAM 
and $123.4$\,ms latency, while on the STM32H7 the footprint drops to 
29{,}492\,B (28.8\,KB) with $94.8$\,ms latency. 
These results demonstrate that our implementation pipeline is reliable across different tasks: for smaller-input tasks such as HAR, it achieves low memory footprint and low latency while still producing predictions identical to the PyTorch baseline.

\begin{table}[h]
    \centering
    \scriptsize
    \caption{Experiments Results}
    \begin{subtable}[t]{0.4\linewidth}
    \centering
    \caption{Ablation study (ESP32S3).}
    \begin{tabular}{lcccc}
    \toprule
    Method  & Peak RAM  & Latency \\
    \midrule
    Unfused KWS\(\phantom{{}^{-}}\) & 1352KB & --\\
    w/o Lifetime\(\phantom{{}^{-}}\) & 611 KB  & $1256.2$\,ms \\
    MambaLM KWS\(\phantom{{}^{-}}\)& 230 KB  & $1133.2$\,ms \\
    \bottomrule
    \end{tabular}
    \label{tab:perf}
    \end{subtable}
    \hfill
    \begin{subtable}[t]{0.55\linewidth}
    \caption{Numerical agreement. (MCU vs. PyTorch)}
    \begin{tabular}{lccc}
    \toprule
    Task & Avg. $L_\infty$ & Avg. mean err. & Worst-case $L_\infty$ \\
    \midrule
    3-class KWS &
    $2.44{\times}10^{-4}$ &
    $2.23{\times}10^{-5}$ &
    $6.52{\times}10^{-4}$ \\
    10-class KWS &
    $4.04{\times}10^{-4}$ &
    $1.98{\times}10^{-5}$ &
    $1.31{\times}10^{-3}$ \\
    6-class HAR &
    $6.50{\times}10^{-5}$ &
    $1.09{\times}10^{-5}$ &
    $9.23{\times}10^{-5}$ \\
    \bottomrule
    \end{tabular}
    \label{tab:num}
    \end{subtable}\\
    \begin{subtable}[t]{0.8\linewidth}
    \centering
    \caption{Resource usage and latency}
    \begin{tabular}{lcccc}
    \toprule
    Platform \& Task & Classes & Peak RAM  & Latency & Flash Size \\
    \midrule
    ESP-KWS & 3  & 235{,}620 Byte (230 KB)  & $1133.2$\,ms &369{,}444 Byte \\
    ESP-KWS & 10 & 235{,}724 Byte (230 KB)  & $1133.6$\,ms &372{,}336 Byte\\
    \midrule
    STM-KWS & 3  & 282{,}932 Byte (276 KB)  & $934.9$\,ms & 305{,}424 Byte\\
    STM-KWS & 10 & 282{,}932 Byte (276 KB) & $964.1$\,ms & 307{,}248 Byte\\
    \midrule
    ESP-HAR & 6 & 44{,}244 Byte (43.2 KB) & $123.42$\, ms & 360{,}740 Byte\\
    STM-HAR & 6 & 29{,}492 Byte (28.8 KB) & $94.79$\, ms & 308{,}720 Byte\\
    \bottomrule
    \end{tabular}
    \end{subtable}
\end{table}



\section{Conclusion}
We presented, to our knowledge, the first deployment of a Mamba-based neural architecture on a resource-constrained microcontroller: \name, along with a complete pipeline for directly porting models from PyTorch to embedded devices. \name is written entirely in C and free of any runtime dependencies, making it lightweight, self-contained, and readily portable across different MCU platforms or embedded hardware without vendor-specific toolchains. 
\name reduced peak RAM usage by 83.0\%, while preserving bit-level fidelity with the PyTorch Mamba layer outputs and achieving 100\% consistency in classification compared with the PyTorch baselines.

\newpage
\bibliographystyle{unsrtnat}
\bibliography{references}

\newpage
\appendix
\section*{Appendix}
\section{Experiment Setup}
\label{experiment}
\subsection{Hardware \& toolchain}
Experiments were conducted on two representative MCU platforms.
\begin{itemize}[leftmargin=*]
    \item \textbf{ESP32S3}: 240\,MHz, 320\,KB RAM, 8\,MB Flash.  
Built with PlatformIO (framework-espidf~5.3.1), CMake~3.16, esptoolpy~4.5.1, 
ninja~1.9, and toolchain-xtensa-esp-elf~13.2.0 (others as in project manifest).     \item \textbf{STM32H7} (Arduino Portenta H7, M7 core): STM32H747XIH6 @ 480\,MHz, 
511.35\,KB RAM, 768\,KB Flash.  
Built with PlatformIO (platform-ststm32~19.3.0), framework-arduino-mbed~4.3.1, 
tool-dfuutil-arduino~1.11.0, and toolchain-gccarmnoneeabi~7.2.1.  
\end{itemize}
For reference implementation and weight export we used Python~3.11.13 and PyTorch~2.6.0 (FP32).
\subsection{Dataset \& Preprocessing}
We evaluate \name on two tasks: \textbf{keyword spotting (KWS)} and \textbf{human activity recognition (HAR)}. 
\begin{itemize}[leftmargin=*]
\item For KWS, we use the Speech Commands v2 dataset~\cite{speechcommandsv2}.
In the 3-class setting, the labels are \textit{yes}, \textit{no}, and an aggregated \textit{\_unknown\_}. 
In the 10-class setting, the labels are \textit{left}, \textit{no}, \textit{off}, \textit{on}, \textit{one}, \textit{right}, \textit{three}, \textit{two}, \textit{yes}, and aggregated \textit{\_unknown\_}. 
Each input is a 0.1 second audio resampled to 16\,kHz and converted to 40-dim log-Mel filterbank features. 

\item For HAR, we use the UCI-HAR dataset, which contains smartphone sensor signals for six human activities~\cite{anguita2013public}.
The raw 561-dimensional features are zero-padded to 570, then reshaped into an input dimension of 57 with sequence length 10, making them suitable for compact model inference on microcontrollers. 
\end{itemize}

\subsection{Model \& Training}
For KWS, the model is configured with 
input\_dim =~\(40\), hidden\_dim =~\(64\), seq\_len =~\(100\), and num\_classes =~\(3\) \& \(10\). 
The architecture is: 
\[
\text{linear projection} \;\rightarrow\; \text{Mamba}(d\_model{=}64) \;\rightarrow\; 
\text{global temporal pooling} \;\rightarrow\; \text{linear classifier}.
\]

For HAR, the model is configured with input\_dim =~\(57\), seq\_len =~\(10\), and six output classes. 
The architecture is same.

Both models are optimized using Adam (lr=$1\mathrm{e}{-3}$, batch size=32). 
The trained PyTorch weights (fp32) serve as the reference for numerical comparison and are exported into C arrays for \name deployment.

\section{Confusion Matrix}
\label{confusion matrix}
\begin{table}[ht]
\centering
\caption{Classification results. (a) Agreement between PyTorch and C predictions
(identical decisions). (b) Ground-truth vs C predictions on the 6k test set.}
\begin{subtable}[t]{0.44\linewidth}
\centering
\caption{PyTorch vs C}
\begin{tabular}{r|ccc}
PyT$\backslash$C & \_unk & no & yes \\\hline
\_unk & 1882 & 0 & 0 \\
no    & 0 & 2148 & 0 \\
yes   & 0 & 0 & 1970 \\
\end{tabular}
\label{tab:ptc}
\end{subtable}\hfill
\begin{subtable}[t]{0.52\linewidth}
\centering
\caption{Ground Truth vs C (Acc.=92.0\%)}
\begin{tabular}{r|ccc}
GT$\backslash$C & \_unk & no & yes \\\hline
\_unk & 1715 & 219 & 66 \\
no    & 75 & 1904 & 21 \\
yes   & 74 & 24 & 1902 \\
\end{tabular}
\end{subtable}
\end{table}

\begin{table}[ht]
\centering
\scriptsize
\caption{Confusion Matrix (PyTorch vs C Prediction,10-classes)}
\begin{tabular}{c|rrrrrrrrrr}
PyT $\backslash$ C & \_unknown\_ & left & no & off & on & one & right & three & two & yes \\
\hline
\_unknown\_ & 602 & 0   & 0   & 0   & 0   & 0   & 0   & 0   & 0   & 0 \\
left        & 0   & 586 & 0   & 0   & 0   & 0   & 0   & 0   & 0   & 0 \\
no          & 0   & 0   & 589 & 0   & 0   & 0   & 0   & 0   & 0   & 0 \\
off         & 0   & 0   & 0   & 653 & 0   & 0   & 0   & 0   & 0   & 0 \\
on          & 0   & 0   & 0   & 0   & 583 & 0   & 0   & 0   & 0   & 0 \\
one         & 0   & 0   & 0   & 0   & 0   & 590 & 0   & 0   & 0   & 0 \\
right       & 0   & 0   & 0   & 0   & 0   & 0   & 598 & 0   & 0   & 0 \\
three       & 0   & 0   & 0   & 0   & 0   & 0   & 0   & 566 & 0   & 0 \\
two         & 0   & 0   & 0   & 0   & 0   & 0   & 0   & 0   & 633 & 0 \\
yes         & 0   & 0   & 0   & 0   & 0   & 0   & 0   & 0   & 0   & 600 \\
\end{tabular}
\end{table}
\begin{table}[htbp]
\centering
\caption{Confusion Matrix (GT vs C Prediction,10-classes, Acc.=92.5\%)}
\scriptsize
\begin{tabular}{c|rrrrrrrrrr}
GT $\backslash$ C & \_unknown\_ & left & no & off & on & one & right & three & two & yes \\
\hline
\_unknown\_ & 468 & 10 & 27 & 31 & 12 & 9 & 8 & 15 & 14 & 6 \\
left        & 7   & 560 & 3  & 9  & 1  & 3 & 11 & 1  & 0  & 5 \\
no          & 29  & 2   & 555 & 2  & 2  & 2 & 0  & 0  & 5  & 3 \\
off         & 6   & 2   & 0  & 577 & 10 & 0 & 0  & 0  & 4  & 1 \\
on          & 14  & 0   & 0  & 30  & 549 & 2 & 0  & 1  & 4  & 0 \\
one         & 11  & 2   & 2  & 0   & 8   & 569 & 4  & 0  & 2  & 2 \\
right       & 14  & 3   & 1  & 4   & 0   & 2 & 571 & 2  & 3  & 0 \\
three       & 28  & 1   & 0  & 0   & 1   & 1 & 2   & 546 & 21 & 0 \\
two         & 17  & 0   & 0  & 0   & 0   & 2 & 2   & 1   & 575 & 3 \\
yes         & 8   & 6   & 1  & 0   & 0   & 0 & 0   & 0   & 5   & 580 \\
\end{tabular}
\end{table}

\begin{table}[htbp]
\centering
\scriptsize
\caption{Confusion Matrix (PyTorch vs C Prediction, HAR, 6-classes)}
\begin{tabular}{c|rrrrrr}
PyTorch $\backslash$ C & WALKING & W\_USTAIRS & W\_DSTAIRS & SITTING & STANDING & LAYING \\
\hline
WALKING            & 544 & 0   & 0   & 0   & 0   & 0   \\
WALKING\_UPSTAIRS  & 0   & 462 & 0   & 0   & 0   & 0   \\
WALKING\_DOWNSTAIRS& 0   & 0   & 402 & 0   & 0   & 0   \\
SITTING            & 0   & 0   & 0   & 415 & 0   & 0   \\
STANDING           & 0   & 0   & 0   & 0   & 608 & 0   \\
LAYING             & 0   & 0   & 0   & 0   & 0   & 516 \\
\end{tabular}
\end{table}

\begin{table}[htbp]
\centering
\scriptsize
\caption{Confusion Matrix (GT vs C Prediction, HAR, 6-classes,Acc.=92.7\%)}
\begin{tabular}{c|rrrrrr}
GT $\backslash$ C & WALKING & W\_USTAIRS & W\_DSTAIRS & SITTING & STANDING & LAYING \\
\hline
WALKING            & 487 & 3   & 6   & 0   & 0   & 0   \\
WALKING\_UPSTAIRS  & 39  & 426 & 6   & 0   & 0   & 0   \\
WALKING\_DOWNSTAIRS& 18  & 12  & 390 & 0   & 0   & 0   \\
SITTING            & 0   & 3   & 0   & 398 & 89  & 1   \\
STANDING           & 0   & 0   & 0   & 17  & 515 & 0   \\
LAYING             & 0   & 18  & 0   & 0   & 4   & 515 \\
\end{tabular}
\end{table}

\end{document}